\pdfoutput=1

\documentclass[11pt]{article}

\usepackage[final]{acl}

\usepackage{times}
\usepackage{latexsym}

\usepackage[T1]{fontenc}

\usepackage[utf8]{inputenc}

\usepackage{microtype}

\usepackage{inconsolata}

\usepackage{graphicx}

\title{Integral Transformer: Denoising Attention, Not Too Much Not Too Little}

\author{Ivan Kobyzev\thanks{Equal contribution}\hspace{3mm}Abbas Ghaddar\footnotemark[1]\hspace{3mm}Dingtao Hu\hspace{3mm}Boxing Chen \\
Huawei Noah’s Ark Lab, Montreal Research Center, Canada\\
{\footnotesize\texttt{\{ivan.kobyzev,abbas.ghaddar,dingtao.hu,boxing.chen\}@huawei.com}}}

\usepackage{amssymb}
\usepackage{pifont}

\newcommand{\R}{\mathbb{R}}
\newcommand{\vq}{\mathbf{Q}}
\newcommand{\vk}{\mathbf{K}}
\newcommand{\vv}{\mathbf{V}}

\newcommand{\mightmention}[1]{}
\newcommand{\problem}[1]{\textcolor{red}{$\star$}}
\newcommand{\answer}[1]{\textcolor{blue}{$\#$}}
\newcommand{\todoreview}[1]{\textcolor{green}{$@$}}

\usepackage{subcaption}
\usepackage{rotating,csquotes}
\usepackage{xcolor}
\usepackage{multirow}
\usepackage{framed}
\usepackage{booktabs}
\usepackage{amsmath}
\usepackage{upgreek}
\usepackage{amsfonts}

\usepackage[most]{tcolorbox}

\newtcbox{\mybox}[1][]{enhanced, colframe=blue, colback=blue!15, 
	frame style={opacity=0.25}, interior style={opacity=0.25}, 
	nobeforeafter, tcbox raise base, shrink tight, extrude by=1mm, #1}

\usepackage{pgfplots}
\usepackage{array}
\usepackage{xcolor}
\usetikzlibrary{shapes.multipart}

\usepackage{subcaption}
\usepackage{rotating,csquotes}
\usepackage{xcolor}
\usepackage{multirow}
\usepackage{framed}
\usepackage{booktabs}
\usepackage{amsmath}
\usepackage{upgreek}
\usepackage{amsfonts}
\usepackage{arydshln}

\begin{document}
\maketitle
\begin{abstract}

Softmax self-attention often assigns disproportionate weight to semantically uninformative tokens such as special tokens and punctuation, a phenomenon known as attention noise. While recent methods like Cog Attention and the Differential Transformer have addressed this by introducing negative attention scores, they risk discarding useful information. In this paper, we propose the Integral Transformer, a novel self-attention mechanism that denoises attention by integrating signals sampled from the logit distribution. Our approach mitigates noise while preserving the contributions of special tokens critical for model performance. 
Extensive experiments demonstrate that our model outperforms vanilla, Cog, and Differential attention variants on well-established knowledge and reasoning language benchmarks. Moreover, our analysis reveals that employing vanilla self-attention in the lower Transformer layers enhances performance and that the Integral Transformer effectively balances attention distributions and reduces rank collapse in upper layers.

\end{abstract}

\section{Introduction}

Self-attention, a core component of the Transformer architecture~\cite{vaswani2017attention}, has remained a dominant component in state-of-the-art language modeling~\cite{dubey2024llama,yang2024qwen2}, computer vision~\cite{rombach2022high,radford2021learning}, and speech recognition~\cite{radford2023robust} models. Consequently, research efforts continue to focus on enhancing the performance~\cite{shen-etal-2019-tensorized, chang-etal-2021-convolutions}, latency~\cite{daoflashattention,shahflashattention}, and memory efficiency~\cite{xiaoefficient,liu2024deepseek} of vanilla self-attention mechanism.

\begin{figure}[!t]  
    \centering
    \includegraphics[width=1.0\columnwidth]{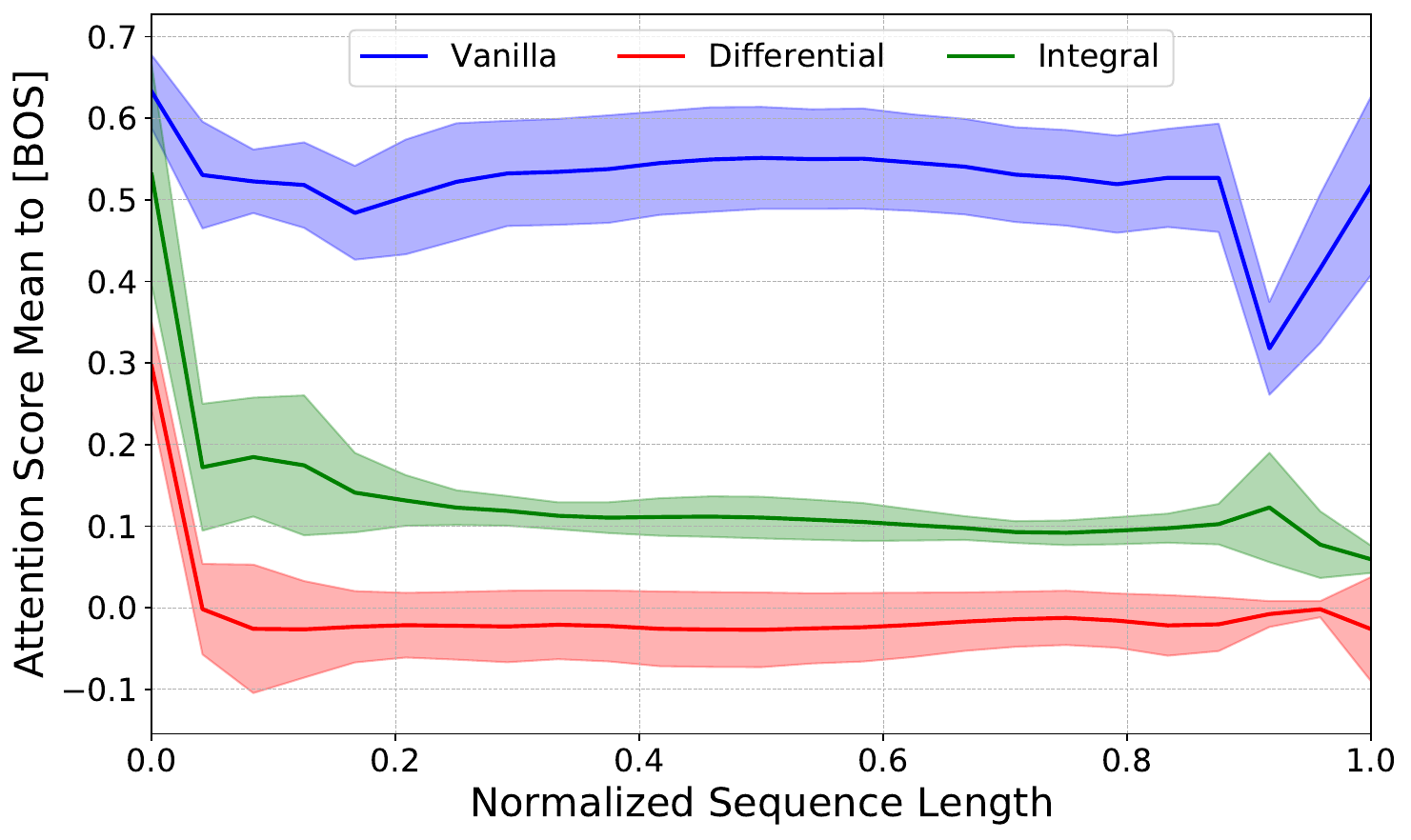}
    \caption{The attention scores (from all tokens in the sequence) in the last layer of Vanilla, Differential, and Integral Transformers with 1.2 billion parameters are measured relative to the beginning-of-sequence \texttt{[BOS]} token. The figure shows the mean (curve) and standard deviation (shaded area) of the attention scores across all attention heads. It is drawn from a randomly selected subset of 1600 samples from the 8 language modeling datasets considered in this study. } 
    \label{fig:bos_attn_main}
\end{figure}

The tendency of Vanilla Transformer Language Models~\cite{arxiv23_llama2} to allocate disproportionately large attention scores to tokens that lack semantic importance (e.g. blue curve in \autoref{fig:bos_attn_main}), such as special tokens or punctuation, has long been a subject of interest~\cite{kovaleva-etal-2019-revealing, clark-2019-bert}. With the rise of large language models (LLMs)~\cite{hurst2024gpt,liu2024deepseek}, this phenomenon has recently attracted attention from machine learning researchers~\cite{sun2024massive,yu2024unveiling}, who have labeled such tokens as non-informative or irrelevant context tokens and often referred to this phenomenon as \textit{attention noise}.

 Cog Attention~\cite{lv2024expressiveattentionnegativeweights} and the Differential Transformer~\cite{ye2024differentialtransformer} propose new self-attention mechanisms that not only diminish the noise but also allow attention scores to become negative for such tokens (red curve in \autoref{fig:bos_attn_main}), thereby reallocating attention to tokens deemed informative and relevant. Although these approaches have shown empirical gains compared to the Vanilla Transformer, they partially contradict findings from both well-established~\cite{clark-2019-bert} and recent~\cite{xiaoefficient,son2024prefixing} literature regarding the role and importance of attending to these tokens.

In this paper, we propose a novel self-attention mechanism that aims to denoise attention by integrating signals sampled from the distribution of logits of the attention layer. We call the resulting model the Integral (\textsc{Intg}) Transformer. Compared to subtracting the scores signals in the Differential (\textsc{Diff}) Transformer, our method still mitigates the noise while retaining attention to special tokens (green curve in \autoref{fig:bos_attn_main}) important for performance. Comprehensive pretraining from scratch experiments demonstrate that our \textsc{Intg} Transformer outperforms both Vanilla, \textsc{Cog}, and \textsc{Diff} Transformers on 8 well-established knowledge and reasoning language evaluation benchmarks.

In addition, we empirically show that maintaining vanilla self-attention in the lower Transformer layers benefits performance. This finding applies not only to our \textsc{Intg} but also to \textsc{Cog} and \textsc{Diff} Transformers. Extensive analysis of the attention head distribution shows that \textsc{Intg} effectively reduces excessive attention to special and punctuation tokens without eliminating it completely, helping to balance the attention score distribution across different token types. Moreover, our analysis reveals that \textsc{Intg} reduces rank collapse~\cite{noci2022signal} in Transformer upper layers more effectively than \textsc{Cog} and \textsc{Diff}.

\color{blue}
\color{black}

\begin{figure*}[!ht]
    \centering
    \includegraphics[width=1.0\textwidth]{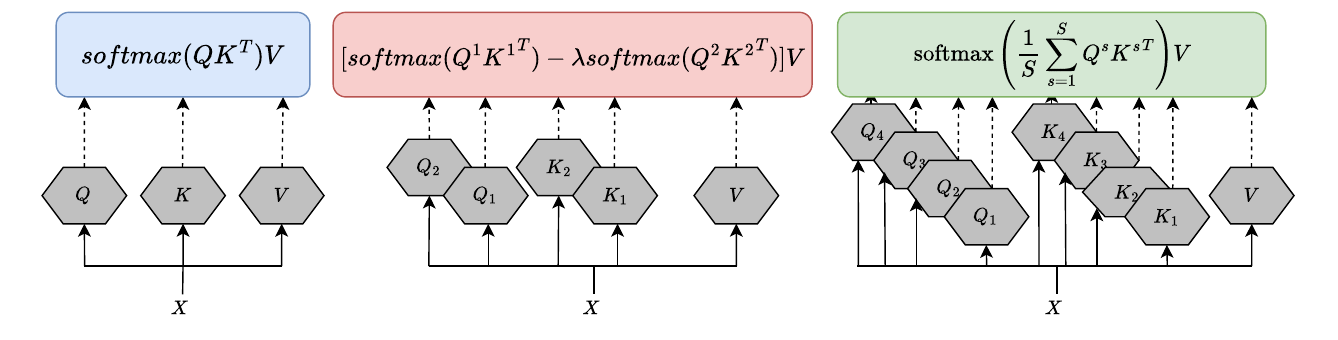}
    \caption{An illustration of the computation of a single attention head representation in the self-attention module of Vanilla (left), Differential (center), and our Integral (right) Transformers, following the notation in \S~\ref{sec:Background} and \S~\ref{sec:Method}. The example illustrates the sampling of 4 signals ($S\!=\!4$) for our Integral Transformer.}
    \label{fig:main_integral}
\end{figure*}

\section{Background}
\label{sec:Background}

We first formulate the self-attention mechanism in the Vanilla Transformer~\cite{vaswani2017attention}, applied to \textit{a single attention head}, in \S~\ref{sec:Self-Attention}. Then, we provide an overview of how the Cog  (\S~\ref{sec:Cog Attention}) and Differential (\S~\ref{sec:Differential Transformer}) methods aim to eliminate attention noise in the vanilla self-attention.

\subsection{Self-Attention}
\label{sec:Self-Attention}

Let $X \in \R^{N\times d_m}$ be a sequence of input representation vectors (e.g., hidden state representations of tokens) for the self-attention module, where $N$ is the length of the sequence and $d_m$ is the dimension of the model. The output of the generalized attention layer is computed as the aggregation of the linear transformation of the input with the attention score which non-linearly depends on the input: 
\begin{align}
\label{eq: SA}
    \text{Self-Attn}(X) = \phi(X)\vv,
\end{align}
where $\vv = XW_V$ for a $d_m \times d_m$ matrix $ W_V$, and $\phi(X) \in \R^{N\times N}$ is the attention score. One of the key properties of the score \(\phi_{ij}(X)\) is that it should capture the degree of relevance of the \(j\)-th token representation, \(X[j,:] \in \mathbb{R}^{d_m}\), with respect to the \(i\)-th token representation.
The original choice for the score computation in self-attention of the \textsc{Vanilla} Transfomer~\cite{vaswani2017attention} is softmax~\cite{Bahdanau2014NeuralMT}:
\begin{align}
    \phi^{o}_W(X) = \text{softmax}(\vq\vk^\top) , \label{eq: Softmax} \\
    \vq = XW_Q, \ \vk = XW_K/\sqrt{d_h}, \label{eq: qk} 
\end{align}
where
$W_Q$ and $W_K$ are ${d_m \times d_h}$ matrices and $d_h$ is a hidden dimension of the score computation (practically, it is a head dimension in a multi-head attention setting). 
\textsc{Vanilla} Transformer with the softmax self-attention block is proven to be a universal approximator of sequence-to-sequence functions \cite{yun2020approximation}. However, this architecture suffers from several representation learning issues like representational collapse~\cite{liu-etal-2020-understanding}, entropy collapse~\cite{krause-entropy} and attention noise~\cite{ye2024differentialtransformer} and the recently proposed modifications of softmax self-attention aims at fixing this.

\subsection{Cog Attention}
\label{sec:Cog Attention}

\citet{lv2024expressiveattentionnegativeweights} proposed a method to increase the flexibility of self-attention by introducing the negative attention scores.
For that, they replaced the softmax operation with the signed softmax:
\begin{align}
\label{eq: Cog}
\scalebox{0.91}{$
    \phi^{cog}(X) = \text{sign}(\vq\vk^\top) \odot
    \text{softmax}(\text{abs}(\vq\vk^\top)), $} 
\end{align}
where $\vq$ and $\vk$ are computed as in Formula~\ref{eq: qk}, $\odot$~is an element-wise product and the functions sign and the absolute value are also applied element-wise.
The authors demonstrate that their approach improves robustness to representational collapse~\cite{noci2022signal}: the phenomenon where token representations at many positions become homogeneous in deeper layers. Moreover, negative weights help eliminate the tendency to focus on non-informative and irrelevant tokens (e.g., special tokens and punctuation).

\subsection{Differential Transformer}
\label{sec:Differential Transformer}

\citet{ye2024differentialtransformer} proposed another approach to mitigate the overallocation of the self-attention to irrelevant context. They replaced the softmax-score computation with a difference of two softmax: 
\begin{align}
\label{eq: Diff}
\begin{split}
    \phi^{dif}(X)  =  \phi^{o}_{W^1}(X)  - \lambda \phi^{o}_{W^2}(X),
    \end{split}
\end{align}
where $\lambda \in \R$ is a learnable parameter (depending on an initialization hyperparameter) and $\phi^{o}(X)$ is the softmax score from Formulas~\ref{eq: Softmax} and \ref{eq: qk} computed with two different sets of $d_m \times d_h$-matrices $W^1_Q$, $W^1_K$ and $W^2_Q$, $W^2_K$. In practice, for efficiency, the hidden dimension of the score computation $d_h$ is taken to be half of the head dimension in a multi-head attention setting.
This approach is based on the differential amplifier technique~\cite{Karki2001FullyDA} from signal processing literature, which primarily calculates the difference between two signals to eliminate common-mode noise in the input. In this context, the authors define this noise primarily as attention to special and less informative tokens.

\section{Method}
\label{sec:Method}

\subsection{Motivation}
\label{sec:Noise}

The tendency of \textsc{Vanilla} Transformer models to allocate significant attention to special or less semantically meaningful tokens, such as punctuation, has been regarded not only as a naturally emerging phenomenon~\cite{clark-2019-bert} but also as important for performance. 
For instance, \citet{han-2024-lm} demonstrates that dropping such tokens during KV-cache compression~\cite{h2o} detrimentally affects LLM performance. \citet{xiaoefficient, oren-2024-transformers} have demonstrated that attending to the first tokens in the sequence (specifically, the \texttt{[BOS]} token) is also crucial for model performance. Furthermore, \citet{dong2024hymba} and \citet{darcet2024vision} found that intentionally adding special tokens at the beginning of the sequence during training phases improves the performance of both language and vision models, respectively. Recently, \citet{zhang2025attention} theoretically studied the importance of these tokens acting as attention sinks for Transformer performance, particularly for few-shot learning~\cite{brown2020language} and chain-of-thought reasoning~\cite{wei2022chain}.

While these studies suggest that it is preferable to keep attending to special tokens, we observe that the approaches in \S~\ref{sec:Cog Attention} and \S~\ref{sec:Differential Transformer} not only remove attention to these tokens but also allow the weights for these tokens to be negative. For instance, we found that 50\% of \textsc{Cog} and 41\% of \textsc{Diff} attention weights to the \texttt{[BOS]} token are negative (see Appendix~\ref{app:Negative Attention Score} for a detailed analysis).
The partial contradiction between the empirical gains obtained by the 
approaches in \S~\ref{sec:Cog Attention} and \S~\ref{sec:Differential Transformer}, and the findings of studies mentioned in the first paragraph of this section, motivate us to propose an alternative method that reduces the noise without completely removing it or allowing the weights for those tokens to go negative.

\subsection{Integral Transformer}
\label{sec:Integral Transformer}

We address attention noise from a different perspective than the differential amplifier technique used by~\cite{ye2024differentialtransformer}, with an approach inspired by spatial antenna diversity from communication system design~\cite{Brennan1959LinearDC}, where signals are diversified to mitigate signal fading and improve the signal-to-noise ratio.

In this approach, we treat logits $\mathbf{Z} = \vq\vk^\top \in \R^{N\times N}$ as signals sampled from some latent distribution: $\mathbf{Z} \sim \mathcal{P}(X)$. We can modify the common-mode noise assumption from the differential transformer. Specifically, we reinterpret noise as zero-mean fluctuations in the logit signals. Then, the natural way to denoise the attention map is to integrate it, which leads to our design of Integral (\textsc{Intg}) Transformer with a new score computation: 
\begin{align}
\label{eq: Integral}
    \phi(X) = \text{softmax}(\mathbb{E}_{\mathcal{P}(X)}[\vq\vk^\top]).
\end{align}

The term \textit{Integral} was adopted by analogy to the \textit{Differential} Transformer, which follows a differential amplifier paradigm. We propose to integrate multiple signals into an attention score function to achieve the attention denoising effect. In practice, we construct the model with an estimation of this integral by averaging. Assuming that $S$ is the number of signals we want to consider, we set up the score computation as a softmax of a signal average: 

\begin{align}
\phi^{intg}(X) = \text{softmax}\left(\frac{1}{S}\sum^{S}_{s=1}\vq^s\vk^{s\top}\right),
\end{align}
\begin{align}
\vq^s = XW^s_Q, \ \vk^s = XW^s_K/\sqrt{d_h}, 
\end{align} 
where $s=1,\dots, S$, $W^s_Q$ and $W^s_K$ are ${d_m \times d_h}$-matrices. It is important to note that this can be implemented in practice without a significant increase in parameters or a loss of efficiency, similar to the original attention mechanism. We do so by choosing the hidden dimension $d_h$ to be the head dimension divided by the number of signals $S$.

\subsection{Signal Design Choice}

In our \textsc{Intg} attention,  we define the signal to be the logits $\mathbf{Z} = \vq\vk^\top \in \R^{N\times N}$, but in differential attention \cite{ye2024differentialtransformer} the signal is taken to be the softmax scores $\phi^o_W(X)$. We will present two theoretical arguments to support our choice. First, integrating signals after the softmax leads to oversmoothing. To demonstrate this, assume that $z$ is an $N$-dimensional Gaussian vector. Then, as  \citet{Shekhovtsov2018FeedforwardUP} show, the expectation of its softmax can be approximated as follows:
\begin{align}\mathbb{E}[\text{softmax}(z)] \approx \text{softmax}(\mathbb{E}[z]/\sqrt{1+\sigma^2}),  
\end{align}
where $\sigma$ is a nontrivial function of the covariance matrix of the Gaussian distribution (see \cite{Shekhovtsov2018FeedforwardUP} for details). This implies that integrating signals after softmax increases the temperature, and hence makes the probability distribution flatter leading to unstable training and worse performance~\cite{anagnostidis2022signal}. 

The second argument concerns the treatment of outliers. The signal after softmax is a categorical probability distribution proportional to the exponential vector: $\text{softmax}(z) \sim \text{exp}(z)$. Averaging the logits corresponds to finding a geometric mean in the score space: $\text{exp}(\frac{1}{S}\sum_sz^s) = \sqrt[S]{\prod_s \text{exp}(z^s)}$. Because taking geometric mean is more resistant to outliers than the arithmetic mean \cite{gupta1982fundamentals}, choosing logits as a signal for the integral attention design is a preferable theoretical choice. We further validate our signal design choice empirically 
in \autoref{tab:sft_abl_app} in Appendix~\ref{app:Results}.

\subsection{Partial Depth Attention Denoising}

Both older and recent studies \cite{clark-2019-bert, xiaoefficient} report different behaviors of deeper and shallower attention layers in their scoring of special tokens. In particular, \citet{xiaoefficient} remarks that lower layers exhibit local attention whereas deeper layers 
 demonstrate increased attention
to initial tokens. 
Additionally, \citet{lv2024expressiveattentionnegativeweights} observe that keeping softmax attention in the first
layer of \textsc{Cog} Transformer significantly enhances the performance. Motivated by findings from prior work on the non-uniformity of the attention mechanism across layers, it is intuitive to question whether applying attention denoising mechanisms to all \textsc{Vanilla} Transformer layers is optimal for performance. In the next section, we attempt to address this question through extensive empirical experiments with a hybrid Transformer model that combines \textsc{Vanilla} and denoising attention layers.

\begin{table*}[!htp]
\centering
\resizebox{\textwidth}{!}{
\begin{tabular}{lccccccccc}

\toprule
\multirow{2}{*}{\textbf{Model}} & \multicolumn{3}{c}{\textbf{Reasoning}} & \multicolumn{5}{c}{\textbf{Knowledge}} \\
\cmidrule(lr){2-4} \cmidrule(lr){5-9}
& \bf Winogrd & \bf ARCe & \bf ARCc & \bf  Hellaswag & \bf PIQA & \bf OBQA & \bf 
 BoolQ & \bf MMLU & \textbf{Avg.}  \\

\midrule
\multicolumn{10}{c}{\textit{125M parameters and 28B tokens}}\\
\midrule 
\textsc{Vanilla}   & 51.3 &  \bf 44.3 & 24.7 & 29.7 & 62.8 & 23.8 & 57.1 & 24.9 & 39.8 \\
\textsc{Cog} & 50.8 & 41.3 & 24.0 & 29.8 & 63.2 & 26.2 & 60.4 & \bf 26.4 & 40.3 \\  
\hspace{5mm}\textit{\small all layers}   & 51.1 & 41.3 & 24.2 & 30.2 & 61.8 & 24.8 & 61.8 & 25.2 & 40.1 \\
\textsc{Diff}  & 52.0 & 41.6 & 24.2 & 29.6 & 63.4 & 25.4 & \bf  62.9 & 24.7 & 40.5 \\     
\hspace{5mm}\textit{\small all layers}   &  \bf 52.3 & 41.9 & 24.3 & 29.9 & 62.3 & 24.8 & 62.1 & 24.9 & 40.3 \\

\textsc{Intg}     & 51.8 & 41.5 &  \bf 26.9 &  \bf 30.4 &  \bf 63.6 &  \bf 28.0 &  62.2 & 24.8 & \bf 41.2 \\
\hspace{5mm}\textit{\small all layers}  & 51.9 & 41.8 & 25.7 & \bf 30.4 & 62.9 & 25.0 & 62.3 & 25.1 & 40.6 \\

\midrule
\multicolumn{10}{c}{\textit{1.2B parameters and 128B tokens}}\\
\midrule
\textsc{Vanilla}                       & 55.6 & 62.0 & 32.0 & 43.7 & 72.8 & 26.0 & 62.1 & 23.1 & 47.2 \\
\textsc{Diff}                          & 54.1 & 62.8 & 32.7 & 43.8 & 73.6 & 26.4 & 62.2 & 25.0 & 47.6 \\
\textsc{Intg}                          &  \bf 56.9 & 62.4 &  \bf 34.3 & \bf  43.9 & \bf  74.8 & \bf  29.8 & 62.2 & \bf  26.9 & \bf 48.9 \\
\hspace{5mm}\textit{\small all layers} & 56.2 &  \bf 63.3 & 33.3 & \bf  43.9 & 73.3 & 28.4 & \bf  62.3 & 24.6 & 48.2 \\

\bottomrule
\end{tabular}
}
\caption{Zero-shot accuracy performance of 4 Transformer architectures, pretrained from scratch under two experimental settings, on 8 language reasoning and knowledge tasks. The main configuration for \textsc{Cog}, \textsc{Diff}, and our \textsc{Intg} (with $S=8$) Transformers consists of these layers in the top 50\% of the Transformer, while the rest are \textsc{Vanilla} Transformer layers. \textit{All layers} indicates results for Transformers when using \textit{all layers} for the aforementioned three architectures. The highest scores for each task under each setting are highlighted in bold.}
\label{tab:main_res}
\end{table*}

\section{Experiments}
\label{sec:Experiments}

\subsection{Experimental Setting}
\label{sec:Experimental Protocol}
We conduct pretraining experiments for LLMs from scratch in two settings: a \textbf{small-scale} setting with 125M parameters and 28B tokens, and a \textbf{large-scale} setting with 1.2B parameters and 128B tokens. We use the Llama2~\cite{arxiv23_llama2} architecture as the backbone in our main experiments in line with prior works on attention noise cancellation~\cite{lv2024expressiveattentionnegativeweights,ye2024differentialtransformer}. We perform standard zero-shot evaluations on eight well-established datasets for commonsense reasoning and knowledge-based language understanding from the LM Eval Harness benchmark~\cite{eval-harness}. A detailed description of the pretraining corpora, implementation details, evaluation datasets, and metrics is available in Appendix~\ref{app:Experimental Setting}. Additionally, we conduct a long-context benchmark evaluation, detailed in Appendix~\ref{app:Long Context Evaluation}.

\subsection{Main Results}

\autoref{tab:main_res} shows the zero-shot accuracy performance of \textsc{Vanilla}~\cite{arxiv23_llama2}, \textsc{Diff}, \textsc{Intg}, and \textsc{Cog} Transformers on 5 knowledge and 3 reasoning language tasks. All models are pretrained from scratch under the two experimental settings described in \S~\ref{sec:Experimental Protocol}. The main configuration for the \textsc{Cog}, \textsc{Diff}, and \textsc{Intg} Transformers involves applying them to only the top 50\% of the layers, while the bottom 50\% use \textsc{Vanilla} Transformer layers. On the small 125M parameter scale, we present ablation results (marked as \textit{all layers}) when \textsc{Cog}, \textsc{Diff}, and \textsc{Intg} are applied to all layers, along with results for the 1.2B scale of our \textsc{Intg} Transformer.

We observe that, at the small scale setting, all three approaches for attention noise cancellation lead to improvements over the \textsc{Vanilla} Transformers by 0.5\%, 0.7\% for \textsc{Cog} and \textsc{Diff}, respectively, with the largest gain of 1.4\% for our \textsc{Intg} on the average across 8 tasks. In addition, we observe a slight yet systematic gain across the three approaches when applying them to the top 50\% of \textsc{Vanilla} Transformers, compared to using all layers.
More precisely, \textsc{Cog} and \textsc{Diff} saw gains of 0.2\%, 0.2\%, and 0.6\%, respectively, compared to their respective \textit{all layers} variants.
Interestingly, despite lagging behind the top 50\% \textsc{Intg}, our \textsc{Intg} \textit{all layers} still outperforms the best \textsc{Cog} and \textsc{Diff} variants, though by a small margin. Moreover, our best \textsc{Intg} variant achieves the highest performance on 4 out of 8 benchmarks when compared against all 5 competing baselines collectively, rather than through one-to-one comparisons. In contrast, each of the remaining baselines ranks highest on at most a single task.

Based on these small-scale performances, we scale up our experiments to 1.2B parameters pretrained on 128B tokens for the \textsc{Vanilla} Transformer, as well as the top three most promising variants under that setting: \textsc{Diff} and both of our \textsc{Intg} variants. Overall, we observe similar trends at large scale compared to small scale, where the best previous architecture, \textsc{Diff}, outperforms \textsc{Vanilla} by 0.4\%, and our \textsc{Intg} reports the best performance of 0.7\%, outperforming its \textit{all layers} variant. More precisely,  our \textsc{Intg} model achieves the best performance on 6 out of 8 benchmarks, with the remaining 2 top scores reported by our all-layers INTG variant. These results highlight the potential of \textsc{Intg} for improving the pretraining of large language models, as well as the importance of applying it (or any equivalent approach) only to the top layers of the model.

\subsection{Full vs. Partial \textsc{Intg} Transformer}
\label{sec:Partial}

We conduct ablation studies on our approach, testing variants that replace \textsc{Intg} layers with \textsc{Vanilla} at different ratios within the Transformer. \autoref{tab:abl_top_main} presents the average scores on 3 Reasoning (Rsn.) and 5 Knowledge (Klg.) tasks, as well as the average across all 8 tasks, using \textsc{Intg} layers at 100\%, 25\%, 50\%, and 75\%, as well as in the bottom 50\% (-50\%). All experiments are conducted in the small-scale 125M parameters setting, and the full results are presented in \autoref{tab:abl_top_app} in Appendix~\ref{app:Results}.  
\vspace{2mm}
\begin{table}[!htp]
    \begin{center}
        \begin{tabular}{l|ccccc}
            \toprule
             & 25\% & 50\% & 75\% & 100\% & -50\% \\
             \midrule
             Rsn. & 38.6 & \bf 39.5 & \underline{38.2} & 39.9 & 38.8 \\
             Klg. & 40.1 & \bf 41.2 & 40.7 & 40.6 & \underline{39.2} \\
             Avg. & 39.6 & \bf 40.6 & 39.7 & 40.4 & \underline{39.0} \\
            \bottomrule  
        \end{tabular}
    \end{center}	
    \caption{Performance of models using \textsc{Intg} in their top 25\%, 50\%, and 75\% of Transformer layers (with the rest being \textsc{Vanilla} layers), as well as the full (100\%) and bottom 50\% (-50\%) \textsc{Intg} Transformer. Bold and underline indicate the highest and lowest scores under the average of reasoning (Rsn.) and knowledge (Klg.) tasks, as well as the overall average across all 8 tasks. Experiments are performed in the small-scale pretraining setting}
    \label{tab:abl_top_main}
\end{table}

First, it is important to note that we use a minimum of 2 signals in these ablation experiments to avoid any potential impact from a mismatch in head size between \textsc{Vanilla} and \textsc{Intg} Transformer layers. Results show that the top 50\% \textsc{Intg} Transformer consistently performs the best across both reasoning and knowledge task categories. Meanwhile, making the entire model consistent by employing \textsc{Intg} across all layers (100\%), is on par but slightly worse than this variant. Interestingly, we notice that applying the \textsc{Intg} Transformer to the bottom 50\% of layers leads to significant underperformance compared to the \textsc{Vanilla} Transformer (first line in \autoref{tab:main_res}), while applying it to the top 25\% or 75\% results in slightly worse performance compared to the \textsc{Vanilla} Transformer.

Finally, it is worth noting that the results show no direct correlation between reducing the effective attention head dimension (due to signal sampling) in \textsc{Intg} Transformer layers and model performance, as the top 25\% and 75\% of models perform roughly the same. These empirical results suggest that \textit{attention noise} is not of the same nature across layers and that not all noise necessarily needs to be canceled for optimal performance. However, deeper and more theoretical studies are needed to fully understand this phenomenon.

\begin{figure*}[t]
    \begin{center}
        \includegraphics[width=\textwidth]{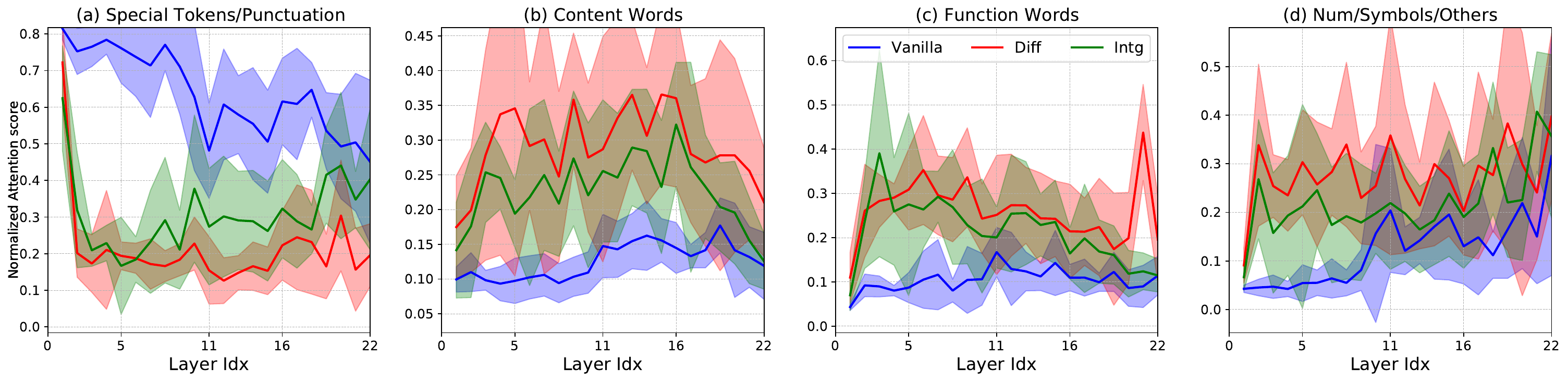} 
    \end{center}	
    \caption{Normalized attention scores of continuation tokens (see Appendix~\ref{app:Benchmark Evaluation}) attending to all tokens in the sequence, grouped into four token types. These scores are computed for all layers of the \textsc{Vanilla}, \textsc{Diff}, and \textsc{Intg} 1.2B-scale Transformer models. The curve represents the mean, and the shaded area indicates the standard deviation across all attention heads.}
    \label{fig:attn_cat_anly}	
\end{figure*}

\subsection{Integral Signals and Heads}
\label{sec:Integral Signals and Heads}

We further study the potential impact on the performance of the hidden dimension $d_h$ 
in the \textsc{Intg} Transformer, exploring different combinations of attention heads and number of signals $S$. \autoref{tab:sig_abl_main} shows the models' (top 50\%) performance~\footnote{Full results are presented in \autoref{tab:sig_abl_app} in Appendix~\ref{app:Results}} when ablating with 2, 4, and 8 signals, using 8 and 16 attention heads in the small-scale 125M parameter and 28B token pretraining setting. We observe a significant increase in performance when scaling up the number of signals from 2 to 8 while using 8 attention heads. This is expected, as sampling more signals provides a better estimation of the  expectation of the signals in Eq.~\ref{eq: Integral}.

In contrast, when doubling the number of heads to 16, we observe a systematic degradation in performance as more signals are sampled, with 16 heads and 8 signals \textsc{Intg} underperforming the \textsc{Vanilla} Transformer. However, it is important to recall that the hidden score size $d_h$ in \textsc{Intg} decreases by a factor equal to the number of signals, which may lead to a very small effective head size. For instance, in a 125M parameter model (hidden model size, $d_m=768$), using 16 heads and 8 signals results in $d_h=6$, which most likely causes poor performance. 

\begin{table}[!ht]
    \begin{center}
    \resizebox{\columnwidth}{!}{
        \begin{tabular}{lccc|ccc}
            \toprule
             & \multicolumn{3}{c}{\textbf{heads=8}} & \multicolumn{3}{c}{\textbf{heads=16}} \\
            \cmidrule(lr){2-4} \cmidrule(lr){5-7}
             \hspace{4mm}$\mathbf{S}$ & $2$ & $4$ & $8$ & $2$ & $4$ & $8$ \\
             \midrule
             Rsn. & 39.5 & 39.8 & 40.1 & \bf 40.7 & 39.4 & 37.7 \\
             Klg. & 41.2 & 41.1 & \bf 41.8 & 41.2 & 41.1 & 40.7 \\
             Avg. & 40.6 & 40.6 & \bf 41.2 & 41.0 & 40.5 & 39.6 \\
            \bottomrule
        \end{tabular}
        }
    \end{center}	
    \caption{Performances of top 50\% \textsc{Intg} models when ablating the number of attention \textbf{heads} and signals ($\mathbf{S}$) values under a small-scale pretraining setting. Bold text indicates the highest scores for each task group. }
    \label{tab:sig_abl_main}
\end{table}

However, increasing the number of heads generally decreases the performance of \textsc{Intg}, which follows the trend observed in the \textsc{Vanilla} Transformer (see \autoref{tab:vanilla_head_app} in Appendix~\ref{app:Results}). This also aligns with \newcite{michel2019sixteen,brown22wide}, who show that adding more heads can degrade performance in the \textsc{Vanilla} Transformer due to redundant information encoding. Therefore, it is important to determine the maximum number of signals that leads to a sufficient head size, while using the same number of heads recommended for a specific \textsc{Vanilla} Transformer configuration.

\subsection{Transformer Backbone Ablation}
\label{sec:Transformer Backbone Ablation}

In this section, we analyze whether the gains of our \textsc{Intg} Transformer are tied to design choices of the Llama2 architecture by experimenting with Pythia~\cite{biderman2023pythia} and Qwen2~\cite{yang2024qwen2} as backbone architectures. 

\begin{table}[!ht]
    \begin{center}
        \resizebox{\columnwidth}{!}{
        \begin{tabular}{lcccccc}
            \toprule
             & \multicolumn{2}{c}{\textbf{LLama2}} & \multicolumn{2}{c}{\textbf{Pythia}} & \multicolumn{2}{c}{\textbf{Qwen2}} \\
            \cmidrule(lr){2-3} \cmidrule(lr){4-5} \cmidrule(lr){6-7}
             & \textsc{Vanl} & \textsc{Intg} & \textsc{Vanl} & \textsc{Intg} & \textsc{Vanl} & \textsc{Intg}\\
             \midrule
             Rsn.  & 40.1 & 39.9 & 39.8 & \bf 41.1 & 40.0 & 40.9 \\
             Klg. & 39.7 & 40.6 & 39.3 & 40.0 & 39.9 & \bf 41.4 \\
             Avg. & 39.8 & 40.4 & 39.5 & 40.3 & 40.0 & \bf 41.0 \\             
            \bottomrule
            
        \end{tabular}
        }
    \end{center}	
    \caption{Performance of the \textsc{Vanilla} and \textsc{Intg} Transformer models using different backbone architectures under a small-scale pretraining setting. Bold text indicates the highest scores for each task group.}
    
    \label{tab:backbone_abl_main}
\end{table}

\autoref{tab:backbone_abl_main} shows the performance of pretraining from scratch under the small-scale setting (125M parameter and 28B token) on reasoning and knowledge language tasks for the \textsc{Vanilla} and \textsc{Intg}~\footnote{We conduct this ablation with \textsc{Intg} when using it at \textit{all layers}. Detailed results are presented in \autoref{tab:backbone_abl_app} in Appendix~\ref{app:Results}} Transformer models. The results show systematic gains for \textsc{Intg} over \textsc{Vanilla} in Pythia (a predecessor to Llama2) and Qwen2 (a slightly enhanced version of Llama2). 
Additionally, we observe that the relative performance of the \textsc{Vanilla} Transformer persists when we apply \textsc{Intg}, suggesting that the surrounding blocks of self-attention modules complement the gains achieved with \textsc{Intg}.

\section{Analysis}
\label{sec:Analysis}

We investigate attention score matrices to understand better how our method shifts the distribution of token types (\S~\ref{sec:Attention Distribution Over Tokens}), the concentration of attention (\S~\ref{sec:Attention Entropy}), and the potential link between attention noise and rank collapse (\S~\ref{sec:Rank Collapse}). The analyses in this section are conducted on a randomly selected subset of 200 samples from each of the 8 tasks, resulting in a total of 1600 samples.
 
\subsection{Token Type Attention Distribution}
\label{sec:Attention Distribution Over Tokens}

We start by analyzing the distribution of attention weights in denoising attention methods across different types of tokens, depending on their degree of informativeness or semantic meaning. To this end, we use a POS tagger~\cite{honnibal2020spacy} to assign a tag to each token in the sequence, with the tags then grouped into the following 4 categories: special tokens (e.g., \texttt{[BOS]}), content words (e.g., nouns), function words (e.g., prepositions), and a category for numbers, symbols, and other tags.\footnote{In the experiment the source consists of continuation tokens only (see Appendix~\ref{app:Benchmark Evaluation}), and the target is all tokens in the sequence. The POS-to-category mapping and other implementation details are listed in Appendix~\ref{app:Token Type Attention Distribution}.}

As illustrated in \autoref{fig:attn_cat_anly}, we conduct this analysis across all 22 layers of the 1.2B \textsc{Vanilla}, \textsc{Diff}, and \textsc{Intg} Transformers. Each sub-figure displays the normalized attention scores for one category (across all layers), with the scores normalized within each layer such that the sum of attention scores across all four categories is $1.0$. The curves represent the mean across all attention heads, while the shaded areas indicate the standard deviation. 

On one hand, we observe that the \textsc{Vanilla} Transformer assigns the highest attention weights to special and punctuation tokens (\autoref{fig:attn_cat_anly} (a)), which is in line with findings from both older and recent studies~\cite{clark-2019-bert,oren-2024-transformers,zhang2025attention}. Additionally, we observe that the \textsc{Vanilla} Transformer assigns the lowest attention weights to the other three token categories (\autoref{fig:attn_cat_anly} (b, c, d)) compared to both the \textsc{Diff} and \textsc{Intg} Transformers. On the other hand, we observe that the \textsc{Diff} model radically reverses these patterns by significantly shifting excessive attention from special and punctuation tokens to the other three categories, which are considered more informative. 

However, we notice that our \textsc{Intg} model also shifts the attention distribution towards more informative tokens, though not as sharply as \textsc{Diff}, as can be clearly seen in \autoref{fig:attn_cat_anly}, where \textsc{Intg} mostly falls in between \textsc{Vanilla} and \textsc{Diff}. These findings, combined with performance gains over \textsc{Vanilla} in \autoref{tab:main_res}, strongly indicate that \textit{attention noise} should not be fully removed, as demonstrated by our \textsc{Intg} Transformer. Finally, we notice that in most cases, \textsc{Diff} and \textsc{Intg} have a significantly higher standard deviation compared to \textsc{Vanilla}. This suggests that the attention heads within the same layer of these two models are not concentrated on the same type of tokens. This finding is further explored in the next section.

\subsection{Attention Concentration}
\label{sec:Attention Entropy}

We study the impact of \textsc{Intg} and other methods on the shift in attention distribution in terms of concentration (spikiness). We do so by measuring the entropy of the attention score distribution for the last token in the continuation segment of the sample sequence.\footnote{See Appendix~\ref{app:Attention Entropy} for implementation details.}

\begin{figure}[!ht]
    \centering
    \includegraphics[width=1.0\columnwidth]{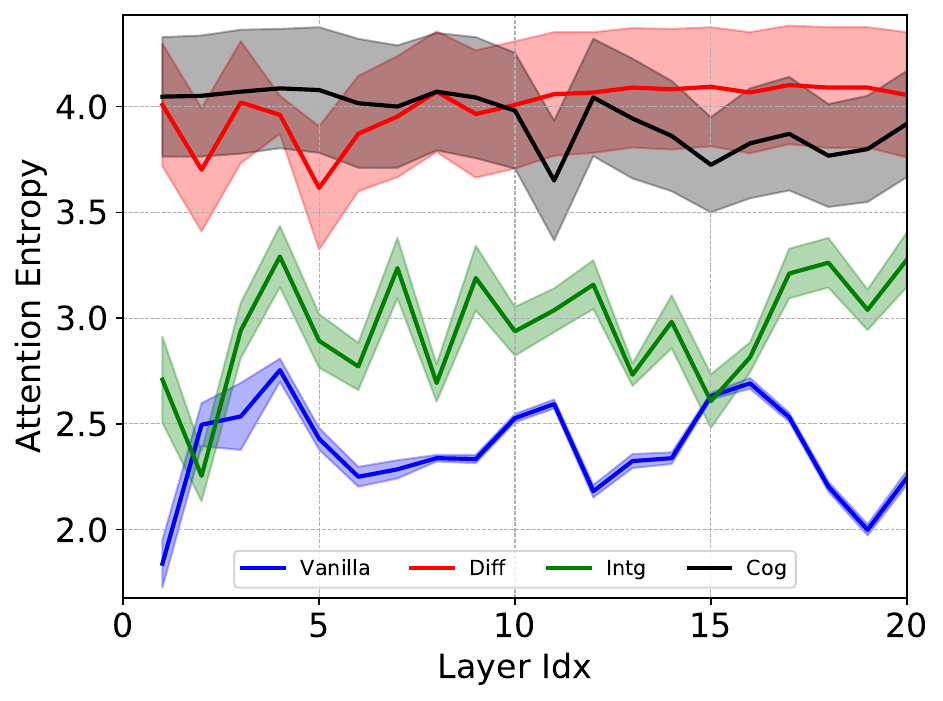}
    \caption{Entropy of the attention score distribution for the last continuation token in each layer of the four 125M parameter Transformer models. The curve represents the mean, and the shaded area indicates the standard deviation over the selected subset of samples.}
    \label{fig:entropy_anly_main}
\end{figure}

On one hand, we notice that \textsc{vanilla} has the lowest entropy, meaning that a large amount of attention is focused on a few tokens. On the other hand, \textsc{Cog} and \textsc{Diff} have the highest overlapping entropy values across all layers, indicating a more uniform dispersion of attention scores over tokens, which the authors of both works intend to achieve with their respective methods. Interestingly, our \textsc{Intg} method stands in the middle at most layers, indicating that it reduces the sparsity of \textsc{Vanilla}, but not as much as the other two methods. This seems to be the most beneficial for performance, as seen in the end-task results in \autoref{tab:main_res}.

\subsection{Rank Collapse}
\label{sec:Rank Collapse}

Rank Collapse in the context of LLMs~\cite{noci2022signal,skean2025layer} refers to a phenomenon where the effective rank of the layer representations gradually decreases as the model deepens. This implies that the model representations become more repetitive or less informative, which is often associated with poor end-task performance. We follow the common approach for analyzing the effective rank of the attention score matrix (\(\phi \in \mathbb{R}^{N \times N}\)) for each attention head in the last four layers of various Transformers, which are more prone to rank collapse. For each data sample, we compute the median rank across all heads and perform a sample-wise comparison between the median rank of \textsc{Vanilla} and each of the \textsc{Cog}, \textsc{Diff}, and \textsc{Intg} Transformer models.   

\begin{table}[!htp]
    \begin{center}
    \resizebox{\columnwidth}{!}{
        \begin{tabular}{l|ccc|ccc}
            \toprule
            & \multicolumn{3}{c}{\textbf{125M}} & \multicolumn{3}{c}{\textbf{1.2B}} \\
            \cmidrule(lr){2-4} \cmidrule(lr){5-7}
             & -1 & -2 & -3 & -1 & -2 & -3  \\
             \midrule
             \textsc{Cog} & 58\% & 70\% & 100\% & - & - & - \\
             \textsc{Diff} & 59\% & 81\% & 100\% & 54\% & 71\% & 63\% \\
             \textsc{Intg} & 69\% & 100\% & 100\% & 62\% & 93\% & 93\% \\
            \bottomrule  
        \end{tabular}
        }
    \end{center}	
    \caption{Average percentage (higher is better) of samples where the rank of an attention-denoising Transformer (row) exceeds that of \textsc{Vanilla} for the 125M and 1.2B parameter scales on the last 3 layers.}
    \label{tab:anly_rank_main}
\end{table}

\autoref{tab:anly_rank_main} shows the average percentage of cases where the rank of an attention-denoising Transformer exceeds that of \textsc{Vanilla} for both the 125M and 1.2B parameter scales. First, we notice that all methods help to mitigate rank collapse compared to \textsc{Vanilla} (all values are $\!>\!50\%$) under both settings. However, \textsc{Intg} reports the best improvements, with 11\% and 10\% higher performance on the 125M scale compared to \textsc{Cog} and \textsc{Diff}, respectively, on the last layer. Secondly, we observe that mitigating rank collapse becomes increasingly challenging as the model depth and size increases, with the rank gap between \textsc{Vanilla} and the denoised attention models narrowing in both settings. These findings suggest that mitigating attention noise is an effective strategy for addressing rank collapse.

\section{Conclusion}
We introduce Integral Transformer, a novel self-attention mechanism that denoises attention by integrating signals sampled from the distribution of the logits. Our approach effectively mitigates attention noise while preserving the influence of special tokens, which are vital for model performance. Experiments on knowledge and reasoning benchmarks demonstrate that the \textsc{Intg} consistently outperforms vanilla and recent alternatives such as \textsc{Cog} and \textsc{Diff} Transformers. 

\section*{Limitations}
Due to limited computational resources, we could not train the model of more than two billion parameters and larger scale and hence, could not properly investigate the scaling law for Integral Transformer. Our evaluation is focused on short-context inputs from an NLP perspective, with an emphasis on attention mechanisms and their treatment of noise and special tokens, and so our method's effectiveness on long-context inputs was not tested. Besides that, the experiments were conducted on a specific set of NLP benchmarks, and additional evaluation on more diverse domains—such as coding and other specialized tasks—could further validate the generalizability of our technique. Future work will aim to address these limitations.

\section*{Acknowledgements}
We thank the anonymous reviewers for their insightful comments.

\bibliography{custom}

\appendix

\clearpage
\section{Experimental Setting}
\label{app:Experimental Setting}
Due to limited computational resources, we define an experimental pretraining protocol that involves conducting most of the ablations at a scale of 125M parameters and a 28B token corpus, while the main experiment is run on 1.2B parameter models and a 128B token corpora.

\subsection{Pretraining Corpora}
We leverage the Cosmopedia v2 (28B tokens) and the deduplicated FineWeb-Edu (220B tokens) subsets of the  SmolLM-Corpus~\cite{benallal2024smollmcorpus} as pretraining data in our experiments. Cosmopedia v2 is a collection of synthetic data generated by prompting Mixtral-8x7B-Instruct-v0.1~\cite{mixtral} to complete textbooks and stories from chunks carefully selected from RefinedWeb~\cite{penedo2023refinedweb} and RedPajama~\cite{weberredpajama}. FineWeb-Edu is a deduplicated version containing high-quality data from educational web pages, filtered from the FineWeb v1 collection~\cite{penedo2024the}. 

It is worth noting that the SmolLM-Corpus has recently been used as a high-quality pretraining resource in the field, for example by Hymba~\cite{dong2024hymba} and Zamba~\cite{glorioso2024zamba}. When referring to pretraining on a 28B token corpus (with 125M parameter models), we use the Cosmopedia v2 corpus. Conversely, when mentioning the use of 128B tokens (with 1.2B parameter models), we refer to a randomly sampled subset of 100B tokens from FineWeb-Edu, added to the 28B tokens from Cosmopedia v2.

\subsection{Model Configurations}

All reported results use the Llama2~\cite{arxiv23_llama2} architecture as the backbone in our main experiments, in line with prior works on attention noise cancellation~\cite{lv2024expressiveattentionnegativeweights,ye2024differentialtransformer}, unless otherwise specified. The 125M configuration, which partially follows Pythia-125M~\cite{biderman2023pythia}, uses a hidden size of 768, an intermediate size of 1155, 8 attention heads, and 20 hidden layers. For the 1.2B configuration, we follow the TinyLLama model~\cite{zhang2024tinyllama}, which has 22 layers, a hidden size of 2048, an intermediate hidden size of 5632, and 32 attention heads.

For both configurations, we tie embeddings to maximize the number of parameters in the encoder and use the Mixtral~\cite{mixtral} tokenizer and its vocabulary (32k tokens), as it is the model used to generate the 28B-token Cosmopedia v2 corpus. When performing ablation with Pythia~\cite{biderman2023pythia} and Qwen2~\cite{yang2024qwen2} as backbone architectures in \S~\ref{sec:Transformer Backbone Ablation}, we use the same number of layers, hidden size, and attention heads as Llama2, while slightly adjusting the intermediate size to ensure the total number of parameters matches 125M.

\subsection{Implementation Details}

For both Cog Attention~\cite{lv2024expressiveattentionnegativeweights} and Differential Transformer~\cite{ye2024differentialtransformer}, we used their respective open-source code for implementation and adhered to their recommended architectural hyperparameter values for initialization, where applicable (e.g., $\lambda$ for \textsc{Diff}). If not otherwise specified, we use 8 signals (the value of $S$ in \S~\ref{sec:Integral Transformer}) in our default experiments for the \textsc{Intg} Transformer for both 125M and 12.B parameters scale. 

While increasing the number of signals ($S$) enhances denoising, but it also an excessively large $S$ reduces per-signal dimension and hence decreases the performance. Analyzing this trade-off is done in ~\autoref{sec:Integral Signals and Heads}. Hyper-parameter selection of $S$ was conducted at the 125M scale and the resulting number of signals (8) was adopted as-is for our main 1.2B experiment. Similarly, the denoising layer ratio was selected based on experiments at the smaller 125M scale, and the same value was directly applied to the main 1.2B-scale experiments. 

Each model is pretrained on a single GPU server that consists of 8 NVIDIA A800 cards with 80GB of memory each. The pre-training code is based on the PyTorch~\cite{paszke2019pytorch} version of the Transformers library~\cite{wolf2020transformers}. For all models, we use the AdamW~\cite{loshchilov2017decoupled} optimizer with a learning rate decay setting the initial learning rate to 3e-4 with 10,000 warm-up steps. 

To speed up the pretraining in our experiments, 
we use mixed-precision training~\citep{fp16}, and Flash Attention 2 library~\cite{shahflashattention}. In addition, we train all models on fully packed sequences of 2048 tokens in length, and set the maximum per-GPU batch size for each model, which is 16 for 125M parameter models and 4 for 1.2B parameter models. We further speed up the training by setting the gradient accumulation step~\footnote{The values were chosen to achieve a total batch size of 2M tokens, as recommended by ~\cite{tangrethinking}.} to 8 and 32 with the 125M and 1.2B parameters models respectively. 

Pretraining experiments approximately took 4 days for the 125M parameter models and 3 weeks for the 1.2B parameter models, respectively. In terms of pre-training dynamics (convergence speed and stability) \textsc{Intg}, Cog, and DIFF behave the same as standard self-attention. This is because they introduce no additional trainable parameters and rely only on lightweight operations, we observe no measurable slowdown in pre-training compared with self-attention.

\subsection{Benchmark Evaluation} 
\label{app:Benchmark Evaluation}

We conduct comprehensive evaluations of the base language models we pretraining from scratch on the following datasets: Winogrande~\cite{sakaguchi2021winogrande}, ARC (Easy and Challenge)~\cite{allenai:arc}, HellaSwag~\cite{zellers2019hellaswag}, PIQA~\cite{bisk2020piqa}, OpenBookQA~\cite{mihaylov2018can}, BoolQ~\cite{clark2019boolq}, and MMLU~\cite{hendrycksmeasuring}. The first three tasks test models' common sense reasoning, while the remaining five tasks assess language understanding knowledge. We perform zero-shot evaluations on all tasks and report the accuracy for each task, as well as the unweighted average score across eight tasks. For visualization purposes in tables, we use the following acronyms to refer to the eight tasks respectively: Winogrd for Winogrande, ARCe for ARC Easy, ARCc for ARC Challenge, Hellaswag for HellaSwag, PIQA, OBQA for OpenBookQA, BoolQ, and MMLU.

A single data sample is constructed by concatenating the following elements: \texttt{[BOS]} \texttt{[INST]} \textit{\{sys prompt\}} \texttt{[/INST]} \textit{\{context\}} \textit{\{continuation\}} \texttt{[EOS]}, where  \texttt{[*]} represents special tokens, while \textit{\{sys prompt\}}, \textit{\{context\}}, and \textit{\{continuation\}} serve as placeholders for different textual segments. Specifically, \textit{\{sys prompt\}} corresponds to a generic system prompt, \textit{\{context\}} represents the task's question/query/prompt context, and \textit{\{continuation\}} denotes the potential answer.

Model evaluation is performed as follows: both the ground truth answer and the distractor options are independently scored by placing each within the \textit{\{continuation\}} placeholder in the prompt described above. Perplexity is then calculated for each sequence, and the answer from the sequence with the lowest perplexity is selected as the predicted response. Accuracy is then computed based on the predicted response, which is a common practice for evaluating language models, particularly those that have not undergone supervised fine-tuning.

\color{blue}

\color{black}

\section{Results}
\label{app:Results}

\subsection{Results Integrity}

We compare the results obtained with our \textsc{Vanilla} Transformer pretrained from scratch with 1.2B parameters and 128B tokens against two off-the-shelf models that have similar properties to our experimental setup. This comparison is made to ensure the validity of our experimental design and, consequently, the integrity of our results. Otherwise said, the purpose of the integrity check is not to compare against off-the-shelf models, but rather to validate the soundness of our design choices within a unique experimental protocol—specifically pretraining at the 1.2B scale using a 128B tokens corpus.

The first model is TinyLLama~\cite{zhang2024tinyllama}, which has exactly the same number of parameters as ours but was trained on 23 times more data. The second is Cosmo~\cite{cosmo}, which is 50\% larger than our model and uses a pretrained corpus that is 50\% larger than ours. It is worth mentioning that the latter model is pretrained on the Cosmopedia v1 dataset~\cite{benallal2024cosmopedia}, which is a 25B subset of the Cosmopedia v2 dataset that we use. However, the authors pretrained their model for 7 epochs to reach a total of 180B tokens. All three models use the same Llama2 Transformer backbone architecture.

\autoref{tab:res_intg_app} shows the results of TinyLLama and Cosmo (directly copied from their respective reports), along with our \textsc{Vanilla} Transformer under the same evaluation protocol of \S~\ref{app:Benchmark Evaluation} (the results are comparable). We notice that, despite using 23 times less data compared to TinyLLama, our model lags behind by only 2.4\% on the 8-task average. Similarly, we are only outperformed by 2.2\% on average compared to the Cosmo model, which is 50\% larger and uses 50\% more training data.
These observations support the validity of our design choices and, consequently, the reliability of our findings.

\color{blue}
\color{black}

\section{Analysis}
\label{app:Analysis}

\subsection{Token Type Attention Distribution}
\label{app:Token Type Attention Distribution}

We use the default Part of Speech (POS) tagger from spaCy~\cite{honnibal2020spacy} to annotate all samples considered in our analysis with universal dependency\footnote{\url{https://universaldependencies.org/u/pos/}} POS tags. These tags are then grouped into four categories based on their semantic and syntactic meaning:

\begin{itemize}
    \item \textbf{Special Tokens/Punctuation} includes all special tokens list in the prompt of~\S~\ref{app:Benchmark Evaluation}, in addition to  \textit{PUNCT} tag.
    \item \textbf{Content Words} includes words of high semantic weight that carry the core meaning of a sentence. The list of tags includes: \textit{NOUN}, \textit{VERB}, \textit{ADJ}, \textit{ADV}, and \textit{PROPN}.

    \item \textbf{Function Words} includes words with structural role that are critical for grammar but low semantic value. The list of tags includes: \textit{DET}, \textit{ADP}, \textit{CCONJ}, \textit{SCONJ}, \textit{PART}, \textit{AUX}, and \textit{PRON}.

    \item \textbf{Num/Symbols/Others} includes number, symbols and other Rare or ambiguous tokens, which are consider as edge cases.  The list of tags includes: \textit{NUM}, \textit{SYM}, \textit{X}, \textit{SPACE}, and \textit{INTJ}.

\end{itemize}

For a given sample, we accumulate the attention scores from the continuation tokens (source) to all tokens in the sequence (target), grouping them according to the four aforementioned categories. We then compute the mean attention score for each head across all continuation tokens in the sample. Afterward, we calculate the mean attention score over all 1,600 samples in the experiment. Finally, we apply softmax normalization across the four categories, ensuring that the attention scores to the four categories sum to one, making them comparable across models. \autoref{fig:attn_cat_anly} shows the mean and standard deviation across the attention heads for the same process applied at each layer.     

\subsection{Attention Concentration}
\label{app:Attention Entropy}
We compute the entropy of the last token in the continuation segment, as it is the most meaningful one that attends to the maximum number of source tokens in the sequence. For each attention head, we compute the entropy of a given sample as follows:
\begin{align}
    E = - \sum_{i=1}^{N} \hat{a}_i \log(\hat{a}_i), 
\end{align}
where $\hat{a} \in \R^N$ is the normalized attention score over the sequence of the last continuation token. Note that we need to normalize the attention because some of the values can be negative for \textsc{Cog} and \textsc{Diff}. 
We apply the normalization as follows: for an unnormalized attention score $a \in \R^N$, we first calculate $a-\min(a)$ and then divide by the sum of its elements. 
After obtaining the entropy for each attention head, we compute the mean to get a single entropy value for each token per layer. \autoref{fig:entropy_anly_main} shows the mean and standard deviation across the selected samples, with the process applied at each layer.

\subsection{Negative Attention Score}
\label{app:Negative Attention Score}

\begin{figure}[!ht]
    \centering
    \includegraphics[width=1.0\columnwidth]{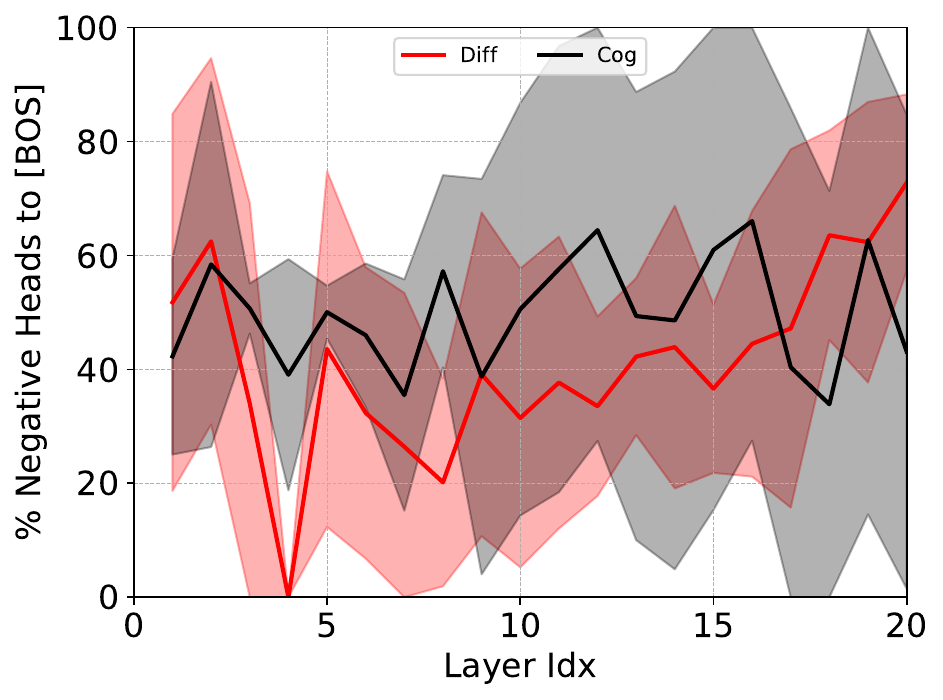}
    \caption{Percentage of attention head scores with negative values from all tokens in the sequence that point to the \texttt{[BOS]} token. These scores are computed for all layers of the \textsc{Cog} and \textsc{Diff} 125M scale Transformer models. The curve represents the mean, and the shaded area indicates the standard deviation across all attention heads.}
    \label{fig:diff_cog_neg_app}
\end{figure}

For each token in a given sample, we compute the percentage of heads with negative attention scores to the \texttt{[BOS]} special token. We then calculate the mean of this percentage across the selected samples that we described at the beginning of~\S~\ref{sec:Analysis} and used there for all analyses. This process is applied to each layer of the small-scale \textsc{Cog} and \textsc{Diff} Transformer models. \autoref{fig:diff_cog_neg_app} shows the mean (curve) and standard deviation (shaded area) over the attention heads of the percentage of heads with negative attention scores to the \texttt{[BOS]}.

Overall, we notice that across all layers (with a few exceptions), the attention scores with negative values are quite significant for both models, generally remaining above 20\% in the majority of cases. It is worth mentioning that, on average across all layers, 50\% and 41\% of the attention heads pointing to \texttt{[BOS]} have negative values for \textsc{Cog} and \textsc{Diff} Transformer, respectively. This observation suggests that these approaches aggressively eliminate attention to \texttt{[BOS]}.    

\subsection{Long Context Evaluation}
\label{app:Long Context Evaluation}
We perform long-context benchmark evaluation of large-scale LLMs on tasks from LongBench~\cite{bai2024longbench} framework. For all models, we perform zero-shot extrapolation of the context length by replacing the default RoPE positional encoding~\cite{su2024roformer} with the Yarn context window extension technique~\cite{pengyarn}. This allows us to achieve meaningful generation at up to 4× the original context length (from 2K to 8K tokens), without any additional finetuning for long-context handling~\cite{wang2024resonance}. 

\begin{table}[!ht]
\centering
\small 
\resizebox{\columnwidth}{!}{
    \begin{tabular}{lccc}

    \toprule
    & \textsc{Vanilla} & \textsc{Diff} & \textsc{INTG} \\
    \midrule
    \multicolumn{4}{c}{\textit{\bf Single-Document QA}} \\
    \midrule
    MultiFieldQA    & \bf 28.26   & 18.71 & 18.66 \\
    NarrativeQA     & 7.84    & 6.11  & \bf 10.76 \\
    Qasper          & 9.20    & 7.99  & \bf 15.97 \\
    \midrule
    \multicolumn{4}{c}{\textit{\bf Multi-Document QA}} \\
    \midrule
    2WikiMQA & \bf 14.27   & 12.96 & 11.69 \\
    HotpotQA & 5.91    & 5.55  & \bf 11.78 \\
    Musique  & \bf 7.96    & 3.42  & 2.32 \\
    \midrule
    \multicolumn{4}{c}{\textit{\bf Summarization}} \\
    \midrule
    GovReport & 8.43    & 4.35  & \bf 10.93 \\
    MultiNews & 6.98    & 6.32  & \bf 8.18  \\
    QMSum     & 8.63    & 11.07 & \bf 15.14 \\
    \midrule
    \multicolumn{4}{c}{\textit{\bf Code Completion}} \\
    \midrule
    LCC         & 9.82    & \bf 15.98 & 15.53 \\
    RepoBench-P & 7.31    & 16.65 & \bf 18.25 \\

    \midrule
    \multicolumn{4}{c}{\textit{\bf Synthetic Tasks}} \\
    \midrule
    Passage Count       & 2.18    & 1.04 & \bf 1.81 \\
    PassageRetrieval & 0.00    & 1.54 & \bf 3.08 \\

    \midrule
    \multicolumn{4}{c}{\textit{\bf Few-shot Learning}} \\
    \midrule
    SAMSum   & 0.45    & 0.86 & \bf 13.71 \\
    TREC     & 8.50    & 2.50 & \bf 11.50 \\
    TriviaQA & 11.07   & 6.95 & \bf 19.26 \\
    
    \bottomrule
    \end{tabular}
    }
    \caption{Long-context benchmark evaluation scores for the \textsc{Vanilla}, \textsc{DIFF}, and our proposed \textsc{Intg} Transformer models across 16 datasets, grouped into 6 tasks from the LongBench benchmark. Bold indicates the highest score for each dataset.  }
    \label{tab:app_long_contx}
\end{table}

Table~\ref{tab:app_long_contx} presents performance (ROUGE scores~\cite{lin2004rouge}) for the 1.2B-scale \textsc{Vanilla}, \textsc{Diff}, and \textsc{INTG} Transformer models on English long-context benchmarks from LongBench, grouped by task type according to the LongBench format\footnote{\url{https://github.com/THUDM/LongBench/tree/main/LongBench}}. Results show that our \textsc{INTG} Transformer outperforms other methods on 12 out of 16 datasets, achieving the highest scores across all datasets in 3 out of the 6 tasks. This observation suggests that attention denoising has the potential to enhance the long-context capabilities of LLMs. However, it worth noting that \textsc{Cog}, \textsc{Diff}, and our \textsc{INTG} methods all focus specifically on addressing noisy attention, without claiming improvements in long-context performance. In fact, \textsc{Cog} did not evaluate on long-context tasks at all, while \textsc{Diff} conducted only synthetic Needle-in-a-Haystack~\footnote{\url{https://github.com/gkamradt/LLMTest_NeedleInAHaystack}} evaluations and only after post-training their models on long context.

\begin{table*}[ht]
\centering
\resizebox{\textwidth}{!}{
\begin{tabular}{l|cc|cccccccc|c}

\toprule
\multirow{2}{*}{\textbf{Model}} & & & \multicolumn{3}{c}{\textbf{Reasoning}} & \multicolumn{5}{c}{\textbf{Knowledge}} \\
\cmidrule(lr){4-6} \cmidrule(lr){7-11}
& \bf \#P & \bf \#T & \bf Winogrd & \bf ARCe & \bf ARCc & \bf  Hellaswag & \bf PIQA & \bf OBQA & \bf 
 BoolQ & \bf MMLU & \bf Avg.  \\  
\midrule
\textbf{TinyLlama} & 1.2B  & 3T  & \bf 59.1 & 55.2 & 30.1 & \bf 59.2 & \bf 73.3 & \bf 36.0 & 57.8 & 25.9  & \bf 49.6 \\
\textbf{cosmo} & 1.8B & 180B          & 54.2 & 56.8 & \bf 33.0 & 55.1 & 71.3 & 35.4 & 53.5 & \bf 32.4 & 49.0 \\ 
\midrule
\bf OurLlama & 1.2B & 128B                     & 55.6 & \bf 62.0 & 32.0 & 43.7 & 72.8 & 26.0 & \bf 62.1 & 23.1 & 47.2 \\
\bottomrule
\end{tabular}
}
\caption{Benchmark evaluation scores of Transformer models, all using Llama2 as the backbone architecture. The models vary in size in terms of the number of parameters (\#P) and are pretrained from scratch on corpora of different sizes, measured in the number of tokens (\#T). The results of TinyLLama~\cite{zhang2024tinyllama} and Cosmo~\cite{cosmo} are taken directly from their respective reports.}
\label{tab:res_intg_app}
\end{table*}

\begin{table*}[ht]
\centering
\resizebox{\textwidth}{!}{
\begin{tabular}{lccccccccc}

\toprule
\multirow{2}{*}{\textbf{Signals}} & \multicolumn{3}{c}{\textbf{Reasoning}} & \multicolumn{5}{c}{\textbf{Knowledge}} \\
\cmidrule(lr){2-4} \cmidrule(lr){5-9}
& \bf Winogrd & \bf ARCe & \bf ARCc & \bf  Hellaswag & \bf PIQA & \bf OBQA & \bf 
 BoolQ & \bf MMLU & \textbf{Avg.}  \\  
\midrule
softmax output & 49.6 & 27.5 & 23.2 & 25.9 & 52.8 & 18.0 & 61.9 & 25.0 & 35.5 \\
logits (our)  & \bf 51.1 & \bf 41.3 & \bf 26.2 & \bf 30.2 & \bf 62.9 & \bf 24.8 & \bf 62.1 & \bf 26.2 & \bf 40.6 \\
\bottomrule
\end{tabular}
}
\caption{Benchmark evaluation performance of \textsc{Intg} Transformer when ablating the signal design choice of either using logits or softmax output as signals. Experiments are run on the 125M parameter Llama2 backbone model and 28B pretraining tokens settings. All models use the \textsc{Intg} Transformer applied to all layers, utilizing 2 signals. \textbf{Bold} shows the highest score for each task. Results clearly justify our theoretical design choice of logits as signals. }
\label{tab:sft_abl_app}
\end{table*}

\begin{table*}[ht]
\centering
\resizebox{\textwidth}{!}{
\begin{tabular}{lccccccccc}

\toprule
\multirow{2}{*}{\textbf{Model}} & \multicolumn{3}{c}{\textbf{Reasoning}} & \multicolumn{5}{c}{\textbf{Knowledge}} \\
\cmidrule(lr){2-4} \cmidrule(lr){5-9}
& \bf Winogrd & \bf ARCe & \bf ARCc & \bf  Hellaswag & \bf PIQA & \bf OBQA & \bf 
 BoolQ & \bf MMLU & \textbf{Avg.}  \\ 
\midrule

\textsc{Intg} & & & &  & & &  & & \\ 
\hspace{3mm} \textbf{top 25\%}    & 50.7 & 40.5 & 24.6 & 30.4 & 62.7 & 22.6 & 60.1 & 24.9 & 39.6 \\
\hspace{3mm} \textbf{top 50\%}    & \bf 51.1 & 41.3 & \bf 26.2 & 30.2 & 62.9 & \bf 24.8 & \bf 62.1 & \bf 26.2 & \bf 40.6 \\
\hspace{3mm} \textbf{top 75\%}    & 50.9 & 40.2 & 23.4 & 30.3 & 62.4 & 24.2 & 61.1 & 25.3 & 39.7 \\
\hspace{3mm} \textbf{top 100\%}  & 51.5 & \bf 42.9 & 25.3 & 30.3 & \bf 63.3 & 23.2 & 61.9 & 24.4 & 40.4 \\
\hspace{3mm} \textbf{bottom 50\%} & 50.1 & 41.6 & 24.6 & \bf 30.4 & 62.5 & 23.6 & 54.7 & 24.7 & 39.0 \\
\bottomrule
\end{tabular}
}
\caption{Benchmark evaluation scores of \textsc{Intg} Transformer models when ablating the percentage of layers where Integral is used, with Vanilla being used otherwise. Experiments are run on the 125M parameter Llama backbone model and 28B pretraining tokens settings with a default number of 8 attention heads and 2 signals. \textbf{Bold} shows the highest score for each task.}
\label{tab:abl_top_app}
\end{table*}

\begin{table*}[ht]
\centering
\resizebox{\textwidth}{!}{
\begin{tabular}{lccccccccc}

\toprule
\multirow{2}{*}{\textbf{Model}} & \multicolumn{3}{c}{\textbf{Reasoning}} & \multicolumn{5}{c}{\textbf{Knowledge}} \\
\cmidrule(lr){2-4} \cmidrule(lr){5-9}
& \bf Winogrd & \bf ARCe & \bf ARCc & \bf  Hellaswag & \bf PIQA & \bf OBQA & \bf 
 BoolQ & \bf MMLU & \textbf{Avg.}  \\  
\midrule

\multicolumn{10}{l}{\textbf{\textsc{Intg} with 8 heads}} \\

\hspace{3mm} signal=2  & 51.1 & 41.3 & 26.2 & 30.2 & 62.9 & 24.8 & 62.1 & \bf 26.2 & 40.6 \\
\hspace{3mm} signal=4  & \bf 53.4 & 41.9 & 24.2 & 30.1 & 62.3 & \bf 28.0 & 61.1 & 24.1 & 40.6 \\
\hspace{3mm} signal=8  & 51.8 & 41.5 & 26.9 & \bf 30.4 & 63.6 & \bf 28.0 & \bf 62.2 & 24.8 & \bf 41.2 \\

\multicolumn{10}{l}{\textbf{\textsc{Intg} with 16 heads}} \\
\hspace{3mm} signal=2  & 50.9 & \bf 43.9 & \bf 27.3 & 30.3 & \bf 63.8 & 25.2 & 60.9 & 25.8 & 41.0 \\
\hspace{3mm} signal=4  & 50.9 & 43.4 & 23.8 & 29.7 & 63.7 & 27.0 & 60.3 & 24.8 & 40.5 \\
\hspace{3mm} signal=8  & 50.0 & 39.8 & 23.4 & 29.0 & 62.3 & 27.4 & 60.0 & 24.9 & 39.6 \\
\bottomrule
\end{tabular}
}
\caption{Benchmark evaluation scores of \textsc{Intg} Transformer models when ablating the number of signals and the number of attention heads. Experiments are run on the 125M parameter Llama2 backbone model and 28B pretraining tokens settings. All models use \textsc{Intg} Transformer layers in the top 50\% and \textsc{Vanilla} Transformer layers in the bottom 50\%. \textbf{Bold} shows the highest score for each task across all settings.}
\label{tab:sig_abl_app}
\end{table*}

\begin{table*}[ht]
\centering
\resizebox{\textwidth}{!}{
\begin{tabular}{lccccccccc}

\toprule
\multirow{2}{*}{\textbf{Model}} & \multicolumn{3}{c}{\textbf{Reasoning}} & \multicolumn{5}{c}{\textbf{Knowledge}} \\
\cmidrule(lr){2-4} \cmidrule(lr){5-9}
& \bf Winogrd & \bf ARCe & \bf ARCc & \bf  Hellaswag & \bf PIQA & \bf OBQA & \bf 
 BoolQ & \bf MMLU & \textbf{Avg.}  \\  
\midrule

\multicolumn{10}{l}{\textbf{Llama2}} \\
\hspace{3mm} heads=4   & 50.3 & 42.4 & \bf24.7 & 29.9 & \bf 63.3 & 22.8 & 59.7 & \bf 25.1 & \bf 39.8 \\
\hspace{3mm} heads=8   & \bf 51.3 & \bf 44.3 & \bf 24.7 & 29.7 & 62.8 & 23.8 & 57.1 & 24.9 & \bf 39.8 \\ 
\hspace{3mm} heads=16  & 51.0 & 39.8 & 23.5 & \bf 30.2 & 62.6 & \bf 24.4 & 57.8 & \bf 25.1 & 39.3 \\ 
\hspace{3mm} heads=32  & 48.2 & 38.7 & 23.1 & 29.0 & 61.7 & 23.0 & \bf 62.1 & 25.0 & 38.8 \\ 
\bottomrule
\end{tabular}
}
\caption{Benchmark evaluation scores of vanilla Transformers (LLaMA 2 backbone) when ablating the number of attention heads. Experiments are conducted on the 125M-parameter LLaMA backbone model with 28B pretraining tokens. Bold indicates the highest score for each task.}
\label{tab:vanilla_head_app}
\end{table*}

\begin{table*}[ht]
\centering
\resizebox{\textwidth}{!}{
\begin{tabular}{lccccccccc}

\toprule
\multirow{2}{*}{\textbf{Model}} & \multicolumn{3}{c}{\textbf{Reasoning}} & \multicolumn{5}{c}{\textbf{Knowledge}} \\
\cmidrule(lr){2-4} \cmidrule(lr){5-9}
& \bf Winogrd & \bf ARCe & \bf ARCc & \bf  Hellaswag & \bf PIQA & \bf OBQA & \bf 
 BoolQ & \bf MMLU & \textbf{Avg.}  \\  
\midrule
\textbf{Llama2}  & 51.3 & \underline{44.3} & 24.7 & 29.7 & 62.8 & \underline{23.8} & 57.1 & \underline{24.9} & 39.8 \\
\hspace{3mm}\textsc{Intg}  & \bf \underline{51.5} & 42.9 & \underline{25.3} & \underline{30.3} & \bf \underline{63.3} & 23.2 & \underline{61.9} & 24.4 & \underline{40.4} \\
\midrule
\textbf{Pythia} & 50.0 & \underline{45.4} & 23.9 & 29.8 & 61.8 & 21.8 & \underline{58.7} & 24.4 & 39.5 \\
\hspace{3mm}\textsc{Intg} & \underline{53.4} & 43.3 & \underline{26.6} & \underline{30.6} & \underline{63.2} & \underline{24.2} & 55.4 & \underline{26.5} & \underline{40.3} \\
\midrule
\textbf{Qwen2} & \underline{50.8} & 43.0 & 26.2 & 29.1 & \underline{63.1} & 24.6 & 59.1 & 24.1 & 40.0 \\
\hspace{3mm}\textsc{Intg} & 50.3 & \bf \underline{45.7} & \bf \underline{26.7} & \bf \underline{31.8} & \underline{63.1} & \bf \underline{26.0} & \bf \underline{59.2} & \bf \underline{26.8} & \bf \underline{41.0} \\ 
\bottomrule
\end{tabular}
}
\caption{Benchmark evaluation scores of Vanilla and our \textsc{Intg} Transformer models when ablating with the backbone model architectures, namely Llama2, Pythia, and Qwen2. Experiments are run on the 125M parameter Llama backbone model and 28B pretraining tokens settings. \textbf{Bold} and \underline{underline} indicate the highest score under all and per-backbone settings, respectively.}
\label{tab:backbone_abl_app}

\end{table*}

\end{document}